\documentclass[letterpaper, 10 pt, conference]{ieeeconf}  

\IEEEoverridecommandlockouts                              

\usepackage{graphics} 
\usepackage{epsfig} 
\usepackage{mathptmx} 
\usepackage{times} 
\usepackage{amsmath} 
\usepackage{amssymb}  
\usepackage{graphicx}
\usepackage{verbatim}
\usepackage{floatrow}

\title{\LARGE \bf Deep Reinforcement Learning for Optimal Critical Care Pain Management with Morphine using  Dueling Double-Deep Q Networks}

\author{Daniel Lopez-Martinez$^{1,2}$, Patrick Eschenfeldt$^{3}$, Sassan Ostvar$^{4}$, Myles Ingram$^{4}$, Chin Hur$^{4}$, and Rosalind Picard$^{2}$
\thanks{$^{1}$Harvard-MIT Health Sciences and Technology
        {\tt\small dlmocdm@mit.edu}}%
\thanks{$^{2}$Massachusetts Institute of Technology (MIT)}%
\thanks{$^{3}$Massachusetts General Hospital}%
\thanks{$^{4}$Columbia University}%
}

\begin{document}

\maketitle

\begin{abstract}
Opioids are the preferred medications for the treatment of pain in the intensive care unit. While under-treatment leads to unrelieved pain and poor clinical outcomes,  excessive use of  opioids puts patients at risk of experiencing multiple adverse effects.
In this work, we present a sequential decision making framework for opioid dosing based on deep reinforcement learning. It provides real-time   clinically interpretable dosing recommendations, personalized according to each patient's evolving pain and physiological condition. We focus on morphine, one of the most commonly prescribed opioids. To train and evaluate the model, we used retrospective data from the publicly available MIMIC-3 database. Our results demonstrate that reinforcement learning  may be used to aid  decision making in the intensive care setting by providing personalized pain management interventions.
\end{abstract}


\section{Introduction}

Critically ill patients in the intensive care unit (ICU) commonly experience pain because of a multitude of factors including surgery, trauma, burns, and cancer, as well as enduring procedural pain \cite{Barr2013}. Regardless of source, poorly managed pain adversely affects multiple organ systems and has a direct impact on outcomes, as the negative physiologic and psychologic consequences of pain are significant, long lasting, and can  preclude patients from participating in their care.

In the ICU, opioids are the primary systemic medications for managing pain. The optimal choice of opioid and the dosing regimen used for an individual patient depend on many factors, including the intensity of pain. To assess the latter as well as to measure  response to treatment, multiple pain intensity scales have been developed. These are unidimensional scales that measure intensity from 0 (no pain) to 10 (worst possible pain).
Unfortunately, pain assessment scales fail in multiple patient populations (e.g. non-verbal children) and clinical situations (e.g. during paralysis). Hence, in recent years, an increasing body of work is addressing the problem of pain measurement using machine leaning with autonomic  \cite{dlmEMBC2018}, brain activity \cite{dlm_ICPR_2018} and facial expression \cite{LopezMartinez2017c} data, and by stratifying patients based on their pain responses  to allow for personalized assessments \cite{dlmNIPS2017}. However, machine learning for automatic pain management interventions remains an unexplored area.

In this work, we propose a novel approach for the automatic dosing of analgesics based on deep reinforcement learning. Specifically, we focus on morphine, one of the most commonly prescribed opioids in the ICU and a first line for the treatment of pain. Both under-treatment and over-treatment with morphine lead to severe adverse effects, such as those related to unrelieved pain and respiratory depression respectively. Our model aims to learn clinically interpretable optimal treatment policies and recommend individualized morphine interventions adapted to each patient's evolving state, to minimize pain while avoiding adverse effects.
To our knowledge, our work is the first to apply reinforcement learning for the recommendation of pain management regimes and the automatic dosing of analgesics.

\section{Background and related work}

Reinforcement learning (RL) algorithms model time-varying state and action spaces, and learn actionable policies $\pi$ (mappings from states to actions) from training examples that do not represent optimal behaviour. At each time step $t$, the RL agent observes the current environment state $s_t$ and takes an action $a_t \in \mathcal{A}$ from a discrete set of actions $\mathcal{A}$, subsequently receiving a reward $r_t$ and transitioning to a new state $s_{t+1}$. The action $a_t$ is chosen following policy $\pi$ so that the return or expected sum of future rewards $R_t = r_t + \gamma r_{t+1} + \gamma^2 t_{t+2}+...$ when taking that action and following policy $\pi$ thereafter is maximized. The future rewards are discounted by a factor $\gamma$ that represents the trade off between  immediate and later rewards.

Combining RL with deep neural networks has achieved tremendous success in complex problems such as the Atari, Go, and chess games \cite{Mnih2015, Silver2018}. In medicine, deep reinforcement learning has been applied, for example, to unfractioned heparin dosing in the ICU \cite{Nemati2016}, intravenous fluid and vasopressor dosing in sepsis patients \cite{Raghu2017,Raghu2017a}, and chemotherapy dosing in cancer patients enrolled in clinical trials \cite{Yauney2018}.

\section{Methods}
Morphine  analgesia is characterized by tremendous interindividual variation and is influenced by multiple factors \cite{Fillingim2005}. Hence, in the ICU, physicians continuously monitor patients to personalize morphine pain management. Our goal is to deduce optimal treatment policies, similar to those of physicians, that account for pain level as well as physiological state as captured by vital signs. 
We chose RL over supervised learning because no ground truth for treatment strategies exists and RL is able to infer optimal strategies from suboptimal training examples.

\subsection{Data and preprocessing}

This work was  conducted on the publicly available Multiparameter Intelligent Monitoring in Intensive Care III (MIMIC-3) database \cite{Johnson2016}, which contains hospital admissions from approximately 38600 individuals aged $\geq 15$ years. From MIMIC-3, we extracted a cohort of ICU patients fulfilling 2 criteria: (1) at least one pain intensity score  had been collected, and (2) at least one dose of intravenous morphine  had been administered.
A total of 6843 admissions fulfilled these criteria. We chose to focus on morphine since this was the most commonly prescribed opioid in the MIMIC-3 database, followed by hydromorphone and oxycodone.

\subsection{Feature extraction}

For each patient, we extracted pain scores, analgesic administration events and vital signs. Data were aggregated into non-overlapping one-hour windows. When several data points were available within a window, we replaced them by either the mean (pain scores, vital signs) or the sum (analgesic administrations).  To account for missing data, we utilized sample-and-hold interpolation as in \cite{Nemati2016}. In total, we extracted 20 hourly features.

\subsubsection{Pain scores}  MIMIC-III  contains scores in both numerical and text format. The latter were converted to a numeric score by using the scores referenced in the text (e.g. \textit{"3-Mild to Mod"} becomes 3). All scores were reported in a 0 (no pain) to 10 (maximum pain) scale. 


\subsubsection{Morphine interventions} 
Morphine sulfate can be administered intravenously either as a bolus  or as a continuous infusion with a steady rate for an extended period of time. In this work, we considered intravenous boluses only, which represented a majority of administrations for the studied admissions (59.4\%). 
These typically have an onset of 5-10 min and a half life of 3-4 hr.  \cite{Narayanan2016a}. The dose, in addition to time of administration, was recorded. 

\subsubsection{Non-morphine analgesic interventions} 
In order to  lower opioid consumption and to improve the quality of analgesia, morphine is commmonly administered with adjuvant analgesics \cite{Narayanan2016a}. This includes nonopiod coanalgesics such as nonsteroidal anti-inflammatory agents. Therefore, we also extracted administrations of 16 analgesics in addition to morphine.


\subsubsection{Physiological signals} 
Morphine induces respiratory depression \cite{Pattinson2008} and impacts the cardiovascular system \cite{Chen2015} with potentially fatal consequences.
Hence, we extracted two relevant physiological signals:  respiration rate (RR) and  heart rate (HR).

\subsection{Continuous State-space and Discretized Action-space }
\label{sec:spaces}
The action space was discretized to 14 actions: $0$, $(0,1]$, $(1,2]$, ..., $(9,10]$, $(10,15]$, $(15,20]$ and $(20,\infty)$ mg of morphine intravenously. The state space, on the other hand, was continuous and directly  captured the patient's self-reported pain intensity and physiological state, as well as non-morphine analgesic interventions. 


\subsection{Reward function}

At each time stamp $t$, the RL agent interacts with the environment and observes the state $s_t$ to  choose whether to deliver a morphine IV bolus (actions 1-13) or to withhold morphine (action 0). The action $a_t$ is chosen to  maximize the return, that is, the expected discounted future reward, defined as $R_t = \sum_{t'=t}^{T} \gamma^{t'-t} r_{t'}$, where $T$ is the terminal timestep.


We defined a reward function $r_t(hr,rr,p)$ that incentivizes 
pain reduction while seeking to maintain heart rate and respiration rate within adequate physiological ranges. For healthy adults at rest, heart rates between 60 and 100 beats per minute and respiration rates between 12 and 20 breaths per minute are considered acceptable. 
\begin{equation}
r_t(hr,rr,p) = C_\text{P} \cdot r^{\text{P}}_t (p_t)
    + C_\text{HR} \cdot  r^{\text{HR}}_t (hr_t)
    + C_\text{RR} \cdot   r^{\text{RR}}_t (rr_t)
\end{equation}
where $C_\text{P} = C_\text{HR} =  C_\text{RR} = \frac{1}{3}$ and 
{\small
\begin{flalign}
    & r^{\text{HR}}_t (hr_t) =     \frac{2}{1 + e^{-({hr}_t- 60)} } 
                        -  \frac{2}{1 + e^{-({hr}_t-100)} } 
                        -  1  &\\
    & r^{\text{RR}}_t (rr_t) =    \frac{2}{1 + e^{-({rr}_t-12)} } 
                       -  \frac{2}{1 + e^{-({rr}_t-20)} } 
                       -  1    &\\ 
    & r^{\text{P}}_t (p_t) = 1 - \frac{2 p_t}{10}  & 
\end{flalign}
}
This function assigns the maximum reward when pain is absent and both heart rate and respiration rate are within the acceptable range, while diminishing as they move away from the ideal range.

\begin{figure}
	\centering
	\includegraphics[width=0.85\linewidth]{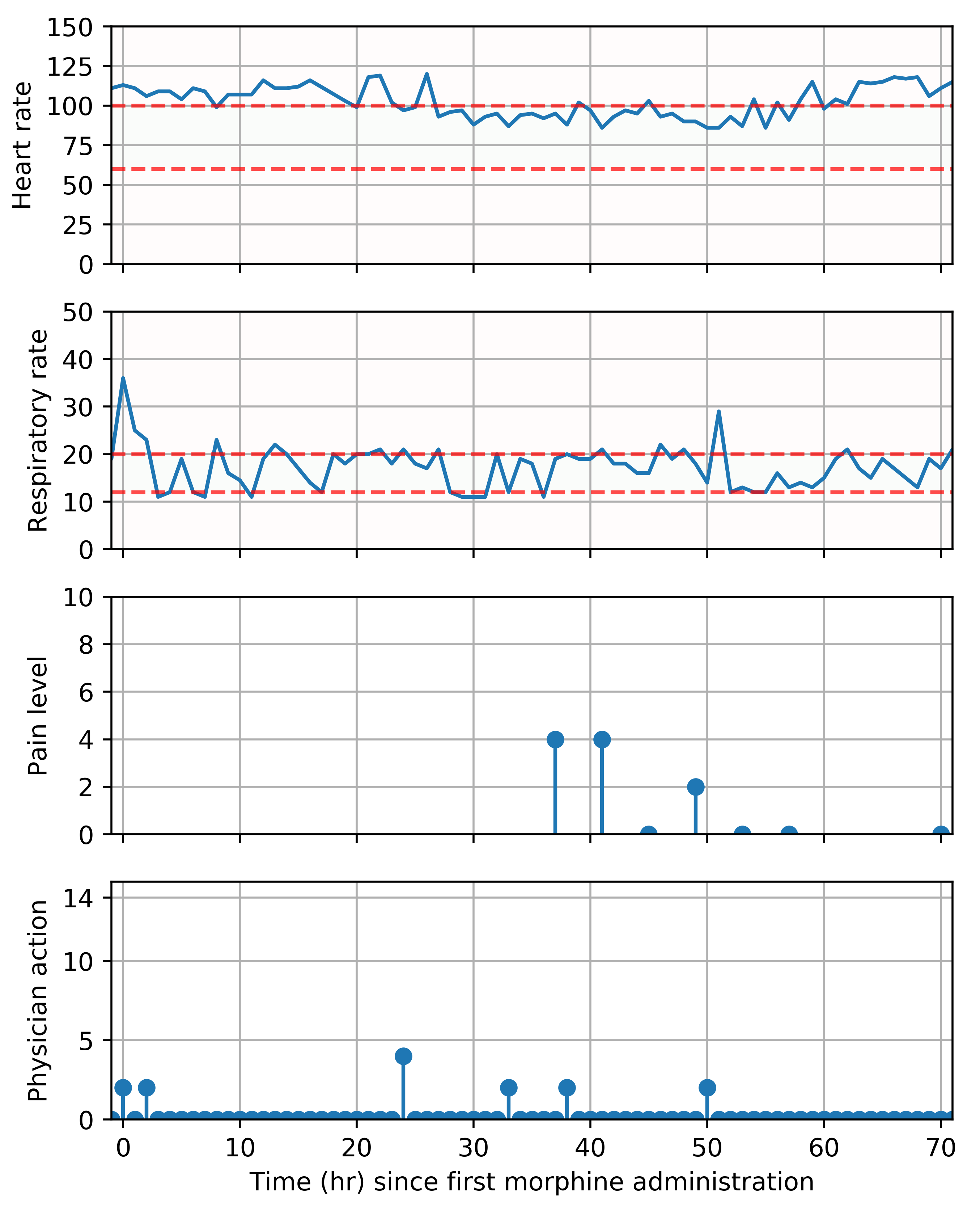}
	\caption{Example of a MIMIC-III patient's time series of physiological data and pain scores, together with the morphine administrations (discretized to 14  actions as described in Sec.\ref{sec:spaces}).}
	\label{fig:Example}
\end{figure}

\subsection{Reinforcement learning model architecture}

To learn the optimal treatment policies, we use Q-learning \cite{Watkins92}, which works by learning an action-value $Q(s_t,a_t)$ function that provides the expected return $R_t$ of taking a given action at each state, that is, 
$
    Q(s_t,a_t) = \max_\pi \mathbb{E}[R_t]
$
where the maximum is taken over all possible policies $\pi$. In Q-learning, the optimal action-value function is estimated using the Bellman equation: 
{ \small
\begin{equation}
Q(s_t, a_t) = \mathbb{E}_{s_{t+1} \sim T(s_{t+1}|s_t,a_t)}  [r + \gamma \max_{a_{t+1}} Q(s_{t+1},a_{t+1}) | s_t, a_t ]
\end{equation}
}
where $T(s_{t+1}|s_t,a_t)$ is the state transition distribution. Deep Q-learning uses neural networks to approximate $Q(s_t,a_t)$ \cite{Mnih2015} by sampling tuples $<s_t,a_t,r_t,s_{t+1}>$ and 
minimizing the squared error loss between the output and  target $Q$ values, where $Q_{\mbox{target}} = r_t+ \gamma \max_{a_{t+1}} Q (s_{t+1}, a_{t+1}; \theta)$ and $\theta$  are the weights used to parametrize the target network, using stochastic gradient descent.

In this work, following \cite{Raghu2017,Raghu2017a}, we combined two common extensions of the deep Q network (DQN) model: double-deep Q networks (DDQN) \cite{Hasselt2016} and dueling Q networks (dueling DQN) \cite{Wang2015}. These address several shortcomings of simple DQNs, such as their tendency to overestimate Q values.

Our final network architecture was therefore a Dueling Double-Deep Q Network (Dueling DDQN), 
It contained two hidden layers with 64 nodes each, with Leaky-ReLu activation functions and equally sized advantage and value streams. As in \cite{Raghu2017, Raghu2017a}, we used prioritized experience replay to sample transitions from the training set with probability proportional to the previous error.  The network was implemented in TensorFlow.

\section{Results}

We separated  6843 hospital admissions into 70\%/20\%/10\% training, validation and test sets respectively. The action space was discretized to 14 actions as described in Sec.\ref{sec:spaces}, based on visual inspection of the histogram of morphine doses in MIMIC-III.
The dueling DDQN model was then trained using Adam optimization \cite{Kingma2015} with batch size 32. After training, we obtained the optimal policy $\pi$ for a given patient state $s$, such that $\pi(s)= \arg \max _a Q(s,a)$. Unfortunately, quantitative evaluation of off-policy models is challenging because it is difficult to assess the impact of deploying the learned policy $\pi$ in patient outcomes. Hence, we evaluated $\pi$ by qualitatively examining the choices of treatments proposed by the model, and comparing these with those made by physicians and our understanding of accepted clinical guidelines.

\begin{figure}
	\centering
	\includegraphics[width=1\linewidth]{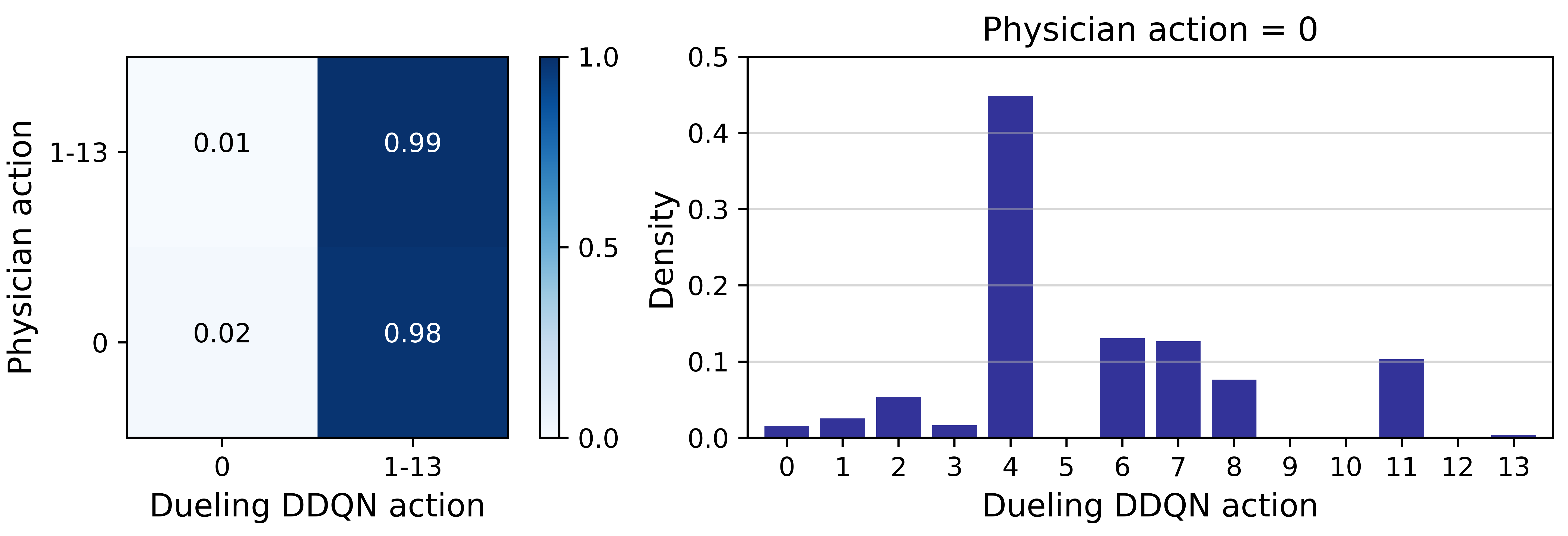}
	\caption{\textit{(Left)} Normalized histogram of physician and dueling DDQN actions, where all actions that involved administration of morphine (actions 1-13) were grouped together. \textit{(Right)} Histogram of dueling DDQN recommended actions for those instances in which physicians chose to withhold morphine (action 0).}
	\label{fig:histogram1}
\end{figure}

\begin{figure}
	\centering
	\includegraphics[width=0.8\linewidth]{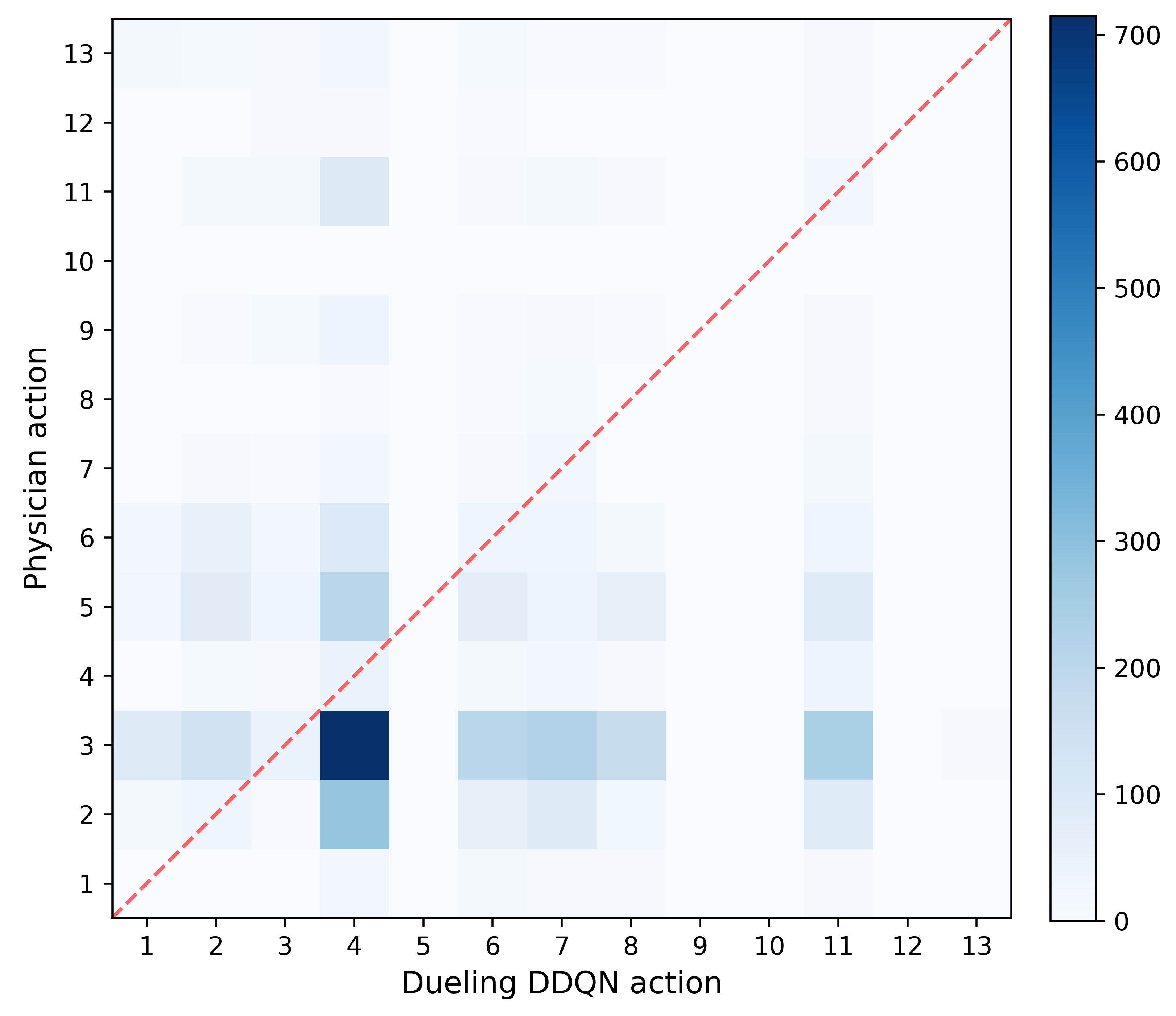}
	\caption{Histogram of physician against dueling DDQN actions  for those instances in which both physician and dueling DDQN recommended  administration of morphine (actions 1-13).}
	\label{fig:2DHistogram}
\end{figure}

Fig.\ref{fig:histogram1} depicts the comparison of actions taken by the physicians with those recommended by the model, for every hour in the test set. Action 0 refers to no morphine administered, whereas actions 1-13 represent increasing morphine doses. Physicians do not often prescribe morphine to patients (only 5.80\% hourly timestamps contained morphine administration), but when they do, in 99\% of instances the model also recommended morphine administration. The agreement between physician actions and model recommendations in these instances is shown in Fig.\ref{fig:2DHistogram}. The actions recommended for the model for the 94.2\% instances in which physicians chose to withhold morphine are shown in Fig.\ref{fig:histogram1}.

\begin{figure}
	\centering
	\includegraphics[width=1\linewidth]{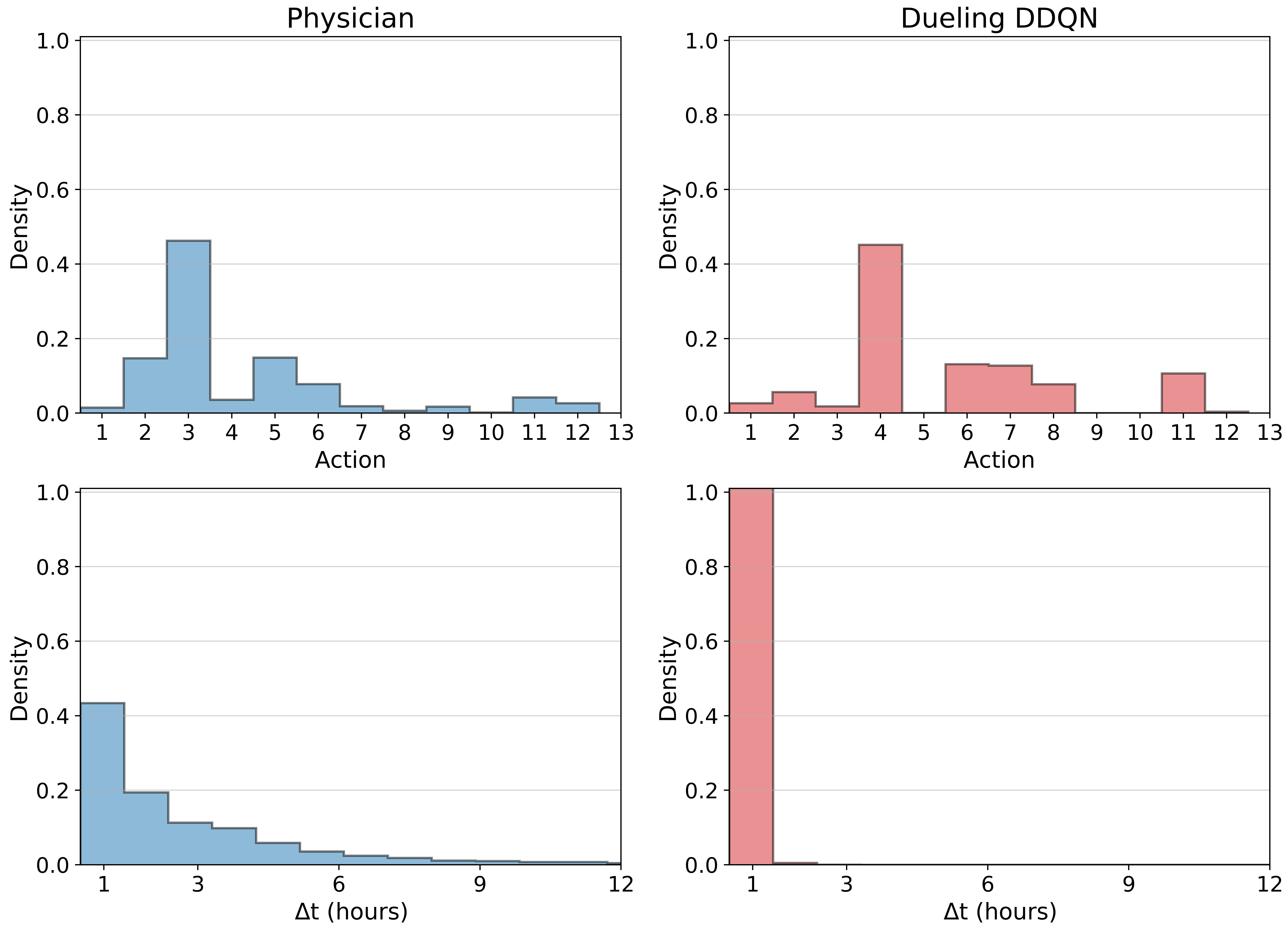}
	\caption{Histogram of morphine administration doses \textit{(top)} and time between administrations \textit{(bottom)} for both physicians \textit{(left)} and the dueling DDQN algorithm \textit{(right)}.}
	\label{fig:1DHistograms}
\end{figure}

While physicians administer morphine sparsely, when they do, a second morphine dose is administered in the next hour in more than 40\% of occasions, as shown in Fig.\ref{fig:1DHistograms}. The model, on the other hand, recommended more continuous administration of morphine. This result is consistent with previous studies comparing intermittent bolus versus continuous infusion dosing of morphine, which have shown that continuous dosing provides similar or even better pain relief with no increase in acute adverse effects \cite{Yu2014,Rutter1980}.

\section{Conclusion}
This work builds on recent research using  reinforcement learning to automatically find optimal treatment policies to improve patient outcome from training examples that do not represent optimal behavior  \cite{Nemati2016,Raghu2017,Raghu2017a}. Here, we focus on the problem of dosing opioid analgesics in the ICU. While machine learning has been applied extensively to the problem of pain recognition, to our knowledge this is the first application of reinforcement learning
for deducing optimal intervention regimens to manage pain. We aimed to demonstrate how deep reinforcement learning could be used to suggest clinically interpretable treatment strategies for pain using real-time patient data that are readily accessible.  
This work has many limitations and should be taken as an illustrative example. Morphine analgesia in the ICU is a complex decision problem influenced by multiple factors. However, our model simplified these clinical factors by constraining  the amount of variables considered 
and defining a reward function that only aims to reduce pain while keeping heart and respiration rates within a pre-defined range. Therefore, possible directions for future work include improving the state space by adding more data modalities, extending the action space to include more dosing regimes (e.g. co-analgesics), and increasing the sophistication of the reward function to reflect the broader range of therapeutic goals. Another limitation is that we used data from a single center, which raises concerns about the generalizability of our findings. Hence, future studies should include multiple centers.
Despite the limitations of this work, our learned policy is consistent with research that shows that continuous dosing is recommended over intermittent dosing. 

Finally, deployment of this approach to opioid dosing in clinical settings would first require extensive prospective evaluation of the learned policies. The goal would not be to replace physicians'  clinical judgments about treatment, but to aid clinical decision making with insights about optimal decisions and automatically guide therapy.







{\small
\bibliographystyle{unsrt} 
\bibliography{Mendeley_v2}

\begin{thebibliography}{10}

\bibitem{Barr2013}
Juliana Barr, Gilles~L. Fraser, Kathleen Puntillo, E.~Wesley Ely, Céline
  G{\'{e}}linas, Joseph~F. Dasta, Judy~E. Davidson, John~W. Devlin, John~P.
  Kress, Aaron~M. Joffe, Douglas~B. Coursin, Daniel~L. Herr, Avery Tung, Bryce
  R.~H. Robinson, Dorrie~K. Fontaine, Michael~A. Ramsay, Richard~R. Riker,
  Curtis~N. Sessler, Brenda Pun, Yoanna Skrobik, and Roman Jaeschke.
\newblock {Clinical Practice Guidelines for the Management of Pain, Agitation,
  and Delirium in Adult Patients in the Intensive Care Unit}.
\newblock {\em Critical Care Medicine}, 41(1):263--306, 1 2013.

\bibitem{dlmEMBC2018}
Daniel Lopez-Martinez and Rosalind Picard.
\newblock {Continuous Pain Intensity Estimation from Autonomic Signals with
  Recurrent Neural Networks}.
\newblock In {\em 2018 40th Annual International Conference of the IEEE
  Engineering in Medicine and Biology Society (EMBC)}, pages 5624--5627,
  Hawaii, 7 2018. IEEE.

\bibitem{dlm_ICPR_2018}
Daniel Lopez-Martinez, Ke~Peng, Sarah Steele, Arielle Lee, David Borsook, and
  Rosalind Picard.
\newblock {Multi-task multiple kernel machines for personalized pain
  recognition from functional near-infrared spectroscopy brain signals}.
\newblock In {\em International Conference on Pattern Recognition (ICPR)},
  Beijing, 2018.

\bibitem{LopezMartinez2017c}
Daniel Lopez-Martinez, Ognjen Rudovic, and Rosalind Picard.
\newblock {Personalized Automatic Estimation of Self-Reported Pain Intensity
  from Facial Expressions}.
\newblock In {\em 2017 IEEE Conference on Computer Vision and Pattern
  Recognition Workshops (CVPRW)}, pages 2318--2327, Hawaii, USA, 7 2017. IEEE.

\bibitem{dlmNIPS2017}
Daniel Lopez-Martinez, Ognjen Rudovic, and Rosalind Picard.
\newblock {Physiological and Behavioral Profiling for Nociceptive Pain
  Estimation Using Personalized Multitask Learning.}
\newblock In {\em Neural Information Processing Systems (NIPS) Workshop on
  Machine Learning for Health}, Long Beach, USA, 2017.

\bibitem{Mnih2015}
Volodymyr Mnih, Koray Kavukcuoglu, David Silver, Andrei~A. Rusu, Joel Veness,
  Marc~G. Bellemare, Alex Graves, Martin Riedmiller, Andreas~K. Fidjeland,
  Georg Ostrovski, Stig Petersen, Charles Beattie, Amir Sadik, Ioannis
  Antonoglou, Helen King, Dharshan Kumaran, Daan Wierstra, Shane Legg, and
  Demis Hassabis.
\newblock {Human-level control through deep reinforcement learning}.
\newblock {\em Nature}, 518(7540):529--533, 2 2015.

\bibitem{Silver2018}
David Silver, Thomas Hubert, Julian Schrittwieser, Ioannis Antonoglou, Matthew
  Lai, Arthur Guez, Marc Lanctot, Laurent Sifre, Dharshan Kumaran, Thore
  Graepel, Timothy Lillicrap, Karen Simonyan, and Demis Hassabis.
\newblock {A general reinforcement learning algorithm that masters chess,
  shogi, and Go through self-play}.
\newblock {\em Science}, 362(6419):1140--1144, 12 2018.

\bibitem{Nemati2016}
Shamim Nemati, Mohammad~M. Ghassemi, and Gari~D. Clifford.
\newblock {Optimal medication dosing from suboptimal clinical examples: A deep
  reinforcement learning approach}.
\newblock In {\em 2016 38th Annual International Conference of the IEEE
  Engineering in Medicine and Biology Society (EMBC)}, pages 2978--2981. IEEE,
  8 2016.

\bibitem{Raghu2017}
Aniruddh Raghu, Matthieu Komorowski, Imran Ahmed, Leo Celi, Peter Szolovits,
  and Marzyeh Ghassemi.
\newblock {Deep Reinforcement Learning for Sepsis Treatment}.
\newblock In {\em Neural Information Processing Systems (NIPS) Workshop on
  Machine Learning for Health}, 2017.

\bibitem{Raghu2017a}
Aniruddh Raghu, Matthieu Komorowski, Leo~Anthony Celi, Peter Szolovits, and
  Marzyeh Ghassemi.
\newblock {Continuous State-Space Models for Optimal Sepsis Treatment - a Deep
  Reinforcement Learning Approach}.
\newblock In {\em Machine Learning for Healthcare Conference}, 2017.

\bibitem{Yauney2018}
Gregory Yauney and Pratik Shah.
\newblock {Reinforcement Learning with Action-Derived Rewards for Chemotherapy
  and Clinical Trial Dosing Regimen Selection}.
\newblock In {\em MLHC}, volume~85, pages 161--226, 2018.

\bibitem{Fillingim2005}
Roger~B. Fillingim, Timothy~J. Ness, Toni~L. Glover, Claudia~M. Campbell,
  Barbara~A. Hastie, Donald~D. Price, and Roland Staud.
\newblock {Morphine responses and experimental pain: Sex differences in side
  effects and cardiovascular responses but not analgesia}.
\newblock {\em Journal of Pain}, 6(2):116--124, 2005.

\bibitem{Johnson2016}
Alistair~E.W. Johnson, Tom~J. Pollard, Lu~Shen, Li~Wei~H. Lehman, Mengling
  Feng, Mohammad Ghassemi, Benjamin Moody, Peter Szolovits, Leo Anthony~Celi,
  and Roger~G. Mark.
\newblock {MIMIC-III, a freely accessible critical care database}.
\newblock {\em Scientific Data}, 3:1--9, 2016.

\bibitem{Narayanan2016a}
Madan Narayanan, A.~Venkataraju, and J.~Jennings.
\newblock {Analgesia in intensive care: part 1}.
\newblock {\em BJA Education}, 16(2):72--78, 2016.

\bibitem{Pattinson2008}
Kyle~T.S. Pattinson.
\newblock {Opioids and the control of respiration}.
\newblock {\em British Journal of Anaesthesia}, 100(6):747--758, 2008.

\bibitem{Chen2015}
Alexander Chen and Michael~A. Ashburn.
\newblock {Cardiac Effects of Opioid Therapy}.
\newblock {\em Pain Medicine (United States)}, 16:S27--S31, 2015.

\bibitem{Watkins92}
Christopher J. C.~H. Watkins and Peter Dayan.
\newblock {Q-learning}.
\newblock {\em Machine Learning}, 8(3-4):279--292, 5 1992.

\bibitem{Hasselt2016}
Hado~Van Hasselt, Arthur Guez, and David Silver.
\newblock {Deep Reinforcement Learning with Double Q-learning}.
\newblock In {\em AAAI'16 Proceedings of the Thirtieth AAAI Conference on
  Artificial Intelligence}, pages 2094--2100, 2016.

\bibitem{Wang2015}
Ziyu Wang, Tom Schaul, Matteo Hessel, Hado van Hasselt, Marc Lanctot, and Nando
  de~Freitas.
\newblock {Dueling Network Architectures for Deep Reinforcement Learning}.
\newblock {\em IEEE Communications Magazine}, 54(1):48--57, 11 2015.

\bibitem{Kingma2015}
Diederik~P. Kingma and Jimmy Ba.
\newblock {Adam: A Method for Stochastic Optimization}.
\newblock In {\em Proceedings of the 3rd International Conference on Learning
  Representations}, 2015.

\bibitem{Yu2014}
Gang Yu, Fu~Qiang Zhang, Shuai~En Tang, Miao~Jun Lai, Rui~Bin Su, and Ze~Hui
  Gong.
\newblock {Continuous infusion versus intermittent bolus dosing of morphine: A
  comparison of analgesia, tolerance, and subsequent voluntary morphine
  intake}.
\newblock {\em Journal of Psychiatric Research}, 59:161--166, 2014.

\bibitem{Rutter1980}
Peter~C. Rutter, F.~Murphy, and H.~A.F. Dudley.
\newblock {Morphine: Controlled trial of different methods of administration
  for postoperative pain relief}.
\newblock {\em British Medical Journal}, 280(6206):12--13, 1980.

\end{thebibliography}
}

\end{document}